\documentclass{article}
\usepackage{ijcai17}
\usepackage[pdftex]{graphicx}
\usepackage{amsmath,amssymb,amsfonts}
\usepackage{subfigure}
\usepackage{xcolor}
\usepackage{algorithm}
\usepackage{algorithmic}
\usepackage{times}
\usepackage{comment}

\newtheorem{Def}{Definition}

\usepackage{hyperref}
\usepackage{color}
\usepackage[normalem]{ulem}
\newcounter{ToDo}
\newcounter{gaocomm}
\newcounter{Note}
\definecolor{blue-violet}{rgb}{0.54, 0.17, 0.89}
\definecolor{mygreen}{rgb}{0.0, 0.5, 0.0}
\definecolor{awesome}{rgb}{1.0, 0.13, 0.32}
\definecolor{bostonuniversityred}{rgb}{0.8, 0.0, 0.0}

\begin{document}
\title{Locality Preserving Projections for Grassmann manifold}
\author{Boyue Wang$^{1}$, Yongli Hu$^{1}$\thanks{Corresponding author: Yongli Hu (huyongli@bjut.edu.cn). Yongli Hu, Yanfeng Sun and Baocai Yin are also with Beijing Advanced Innovation Center for Future Internet Technology.}, Junbin Gao$^{2}$, Yanfeng Sun$^{1}$, Haoran Chen$^{1}$ and Baocai Yin$^{3,1}$\\
$^{1}$Beijing Key Laboratory of Multimedia and Intelligent Software Technology, \\
Faculty of Information Technology, Beijing University of Technology, China\\
$^2$Discipline of Business Analytics. The University of Sydney Business School, University of Sydney, Australia\\
$^3$Faculty of Electronic Information and Electrical Engineering, Dalian University of Technology, China \\
}

\maketitle

\begin{abstract}
Learning on Grassmann manifold has become popular in many computer vision tasks, with the strong capability to extract discriminative information for imagesets and videos. However, such learning algorithms particularly on high-dimensional Grassmann manifold always involve with significantly high computational cost, which seriously limits the applicability of learning on Grassmann manifold in more wide areas. In this research, we propose an unsupervised dimensionality reduction algorithm on Grassmann manifold based on the Locality Preserving Projections (LPP) criterion. LPP is a commonly used dimensionality reduction algorithm for vector-valued data, aiming to preserve local structure of data in the dimension-reduced space. The strategy is to construct a mapping from higher dimensional Grassmann manifold into the one in a relative low-dimensional with more discriminative capability.
The proposed method can be optimized as a basic eigenvalue problem. The performance of our proposed method is assessed on several classification and clustering tasks and the experimental results show its clear advantages over other Grassmann based algorithms.
\end{abstract}

\section{Introduction}

Dimensionality reduction (DR), which extracts a small number of features from original data by removing redundant information and noise, can improve efficiency and accuracy in a wide range of applications, involving facial recognition \cite{HeCaiYanZhang2005,XieTaoWei2016}, feature extraction \cite{LuoNieChangYang2016,WangGao2016} and so on. 
The classic DR algorithms include Locality Preserving Projections (LPP) \cite{HeCaiYanZhang2005}, Principal Components Analysis (PCA) \cite{Bishop2006}, Canonical Correlation Analysis (CCA) \cite{SunCeranYe2010} and Independent Component Analysis (ICA) \cite{Comon1994}.
Most existing DR algorithms are mainly designed to work with vector-valued data, which cannot be directly applied on multi-dimensional data or structured data (i.e., matrices, tensors). 
Simply vectorizing such structured data to fit vector-based DR algorithms  may destroy valuable structural and/or spatial information hidden in data. Therefore, how to effectively and properly reduce the dimensionality of structured data becomes an urgent issue in the big data era.

\begin{figure}
    \begin{center}
    \includegraphics[width=0.5\textwidth]{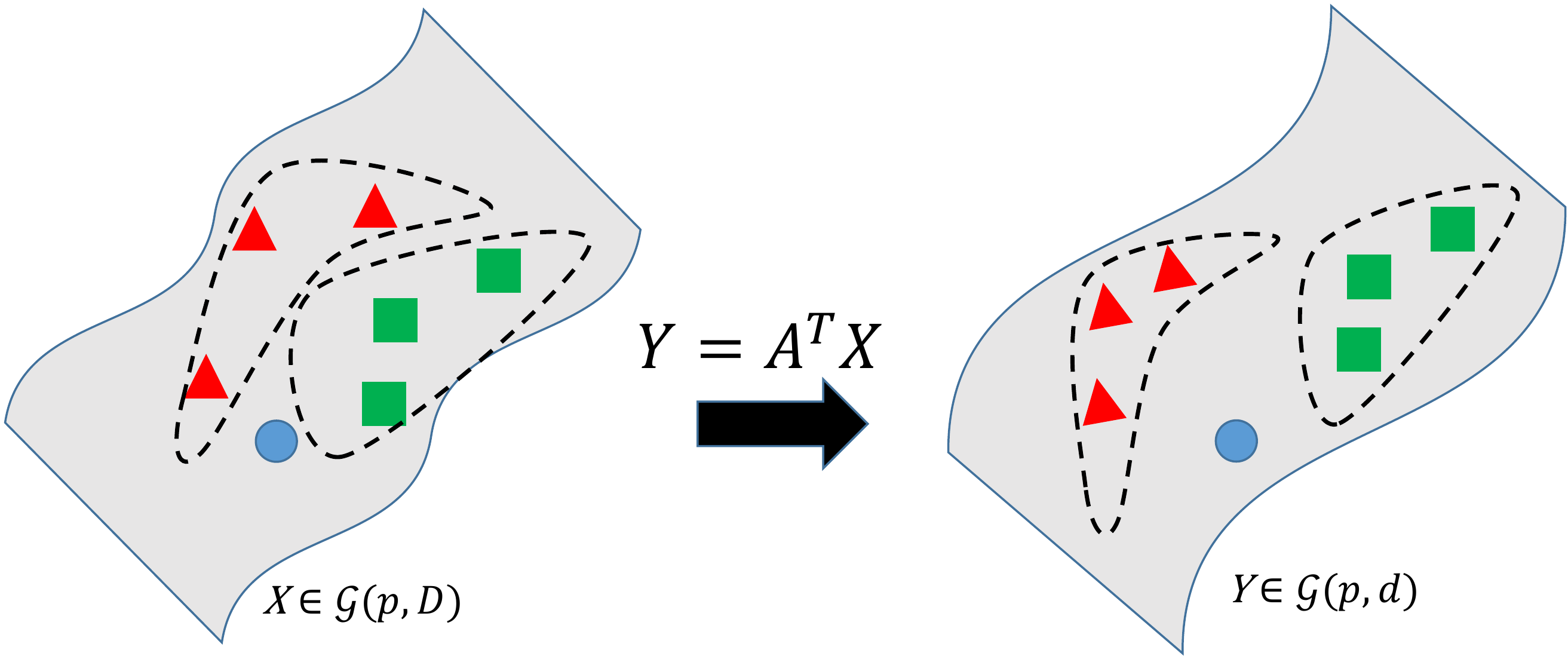}
    \end{center}
    \caption{Conceptual illustration of the proposed unsupervised DR on Grassmann Manifold. 
The Projected Grassmann points still preserve the local structure of original high-dimensional Grassmann manifold.}\label{DRfig}
\end{figure}

In practical application tasks such as those in computer vision, except for well-structured data like matrices or tensors, there exist data which are manifold-valued. For example, in computer vision, the movement of scattered keypoints in images can be described by subspaces, i.e., the points on the so-called Grassmann manifold \cite{AbsilMahonySepulchre2008}; 
and the covariance feature descriptors of images are SPD manifold-valued data \cite{PennecFillardAyache2006}. 
How to design learning algorithms for these two types of special manifold-valued data has attracted great attention in the past two decades \cite{HuangWangShanChen2014,FarakiHarandiPorikli2015b,JayasumanaHartleySalzmannLiHarandi2015}. 
For our purpose in this paper, we will briefly review some recent progress about DR algorithms for structured and manifold-valued data.


For a clear outline, we start with PCA. PCA is the most commonly used DR algorithm for vectorial data. The basic idea of PCA is to find a linear DR mapping such that as much variance in dataset as possible is retained. The classic PCA has been extended to process two dimensional data (matrices) directly with great success \cite{YangZhangFrangiYang2004,YuBiYe2008} (2DPCA). Wang \emph{et al.} \shortcite{WangChenHuLuo2008}  consider the probabilistic 2DPCA algorithm including the algorithm for the mixture of local probabilistic 2DPCA. To identify outliers in structured data, Ju \emph{et al.} \shortcite{JuSunGaoHuYin2015} introduce the Laplacian distribution into the probabilistic 2DPCA algorithm.

Contrary to the global variance constraint in PCA-alike algorithms, LPP focuses on
preserving the local structure of original data in the dimension-reduced space. The first work of extending LPP for 2D data was proposed in
\cite{ChenZhaoKongLuo2007}, which is operated directly on image matrices. The experimental results show that 2DLPP performs better than 2DPCA and LPP. Xu \emph{et al.} \shortcite{XuFengZhao2009} propose a supervised 2DLPP by constructing a discriminative graph of labeled data. To reduce the high computational cost of 2DLPP, Nyuyen \emph{et al.} \shortcite{NguyenLiuVenkatesh2008} improve 2DLPP by using the ridge regression.

However, the aforementioned DR algorithms for matrices are concerned in terms of Euclidean alike distance. 
Although the Riemannian structure has been shown to overcome the limitations of Euclidean geometry of data \cite{PennecFillardAyache2006,HammLee2008}, 
the computational cost of the resulting techniques increases substantially with the increasing dimensionality of manifolds (i.e., the dimension of its embedding space). To the best of our knowledge, few attention has been paid on DR for Riemannian manifold.

Harandi \emph{et al.} \shortcite{HarandiSalzmannHartley2014} extend PCA onto SPD manifold by employing its Riemannian metrics, and then incorporate a discriminative graph of the labeled manifold data to achieve a supervised DR algorithm for SPD manifold. Recent research has shown that the Grassmann manifold, another type of Riemannian matrix manifold, is a good tool to represent videos or imagesets \cite{WangGuoDavisDai2012,HarandiSandersonShenLovell2013,HarandiHartleyLovellSanderson2015,WangHuGaoSunYin2016}.
In a newly proposed supervised metric learning on the Grassmann manifold \cite{HuangWangShanChen2015}, an orthogonal matrix that maps the original Grassmann manifold into a more discriminative one is learned from data.
In handling Grassmann-valued data, one usually employs one of three ways: embedding into a Hilbert feature space defined a Grassmann kernel function \cite{HarandiSandersonShiraziLovell2011}; or embedding into the symmetric matrix manifold (a plain Euclidean space) \cite{WangHuGaoSunYin2016}; or projecting data onto tangent spaces (extrinsic way) \cite{HarandiSandersonShenLovell2013}. However the performance of all these ways can be hindered by the high dimensionality of given Grassmann manifold. It has become critical to reduce the dimensionality of Grassmann data.


Motivated by \cite{HuangWangShanChen2015}, we learn a projected matrix to reduce the dimensionality of Grassmann manifold in this paper. To fulfill the goal, we extend LPP local criterion onto Grassmann manifold through embedding Grassmann manifold into a symmetric matrices space \cite{HarandiSandersonShenLovell2013} such that the local structure of original Grassmann data can be well preserved in the newly projected Grassmann manifold. 
Figure \ref{DRfig} illustrates that a projected matrix $\mathbf A$ is introduced to map the original high-dimensional Grassmann manifold into the one in a relative low-dimensional with more discriminative capability, which still preserves the structure of original Grassmann points.

The contribution of this paper is summarized as follows,
\begin{itemize}
  \item A novel unsupervised DR algorithm in the context of Grassmann manifold is introduced. The DR is implemented by learning a mapping to a Grassmann manifold in a relative low-dimensional with more discriminative capability;
  \item The proposed method generalizes the classic LPP framework to non-Euclidean Grassmann manifolds and 
only involves the basic eigenvalue problem; and
\end{itemize}

We briefly review some necessary knowledge about LPP and Grassmann manifold in next section.

\section{Backgrounds}

\subsection{Locality Preserving Projections (LPP)}

LPP uses a penalty regularization to preserve the local structure of data in the new projected space. 

\begin{Def}[Locality Preserving Projections] \cite{HeNiyogi2003} Let $\mathbf X = [\mathbf x_1, ..., \mathbf x_N]\in\mathbb{R}^{D\times N}$ be the data matrix 
with $N$ the number of samples and $D$ the dimension of data. Given a local similarity $\mathbf W = [w_{ij}]$ among data $\mathbf X$, LPP seeks for the projection vector $\mathbf a$ such that the projected value $y_i = \mathbf a^T\mathbf x_i$ ($i=1, ..., N$)  fulfills the following objective,
\begin{equation}\label{LPP_objective}
\begin{aligned}
\min\limits_{\mathbf a}\sum\limits_{i,j=1}^N(\mathbf a^T\mathbf x_i - \mathbf a^T\mathbf x_j)^2 \cdot w_{ij} = \sum\limits_{i,j=1}^N  \mathbf a^T\mathbf X\mathbf L\mathbf X^T\mathbf a,
\end{aligned}
\end{equation}
with the constraint condition,
\begin{equation}\label{LPP_Constrain}
\begin{aligned}
 \mathbf y\mathbf D \mathbf y^T=\mathbf a^T\mathbf X\mathbf D\mathbf X^T\mathbf a = 1,
\end{aligned}
\end{equation}
where $\mathbf y = [y_1, ..., y_N]$, $\mathbf L = \mathbf D - \mathbf W$ is the graph Laplacian matrix and $\mathbf D = \text{diag}[d_{ii}]$ with $d_{ii}=\sum\limits_{j=1}^N w_{ij}$. 
\end{Def}

A possible definition of $\mathbf W$ is suggested as follows:
\begin{equation}
w_{ij} =
\begin{cases}
e^{-\frac{\|\mathbf x_i - \mathbf x_j\|^2}{t}}, & \ \ \mathbf x_i \in \mathcal{N}(\mathbf x_j) \ \ \text{or} \ \ \mathbf x_j \in \mathcal{N}(\mathbf x_i); \\
0,& \ \ \text{otherwise}. \\
\end{cases}
\end{equation}
where $t \in \mathbb{R}_+$ and $\mathcal{N}(\mathbf x_i)$ denotes the $k$ nearest neighbors of $\mathbf x_i$. With the help of $\mathbf W$, minimizing LPP objective function \eqref{LPP_objective} is to ensure if $\mathbf x_i$ and $\mathbf x_j$ are similar to each other, then the projected values $y_i = \mathbf a^T \mathbf x_i$ and $y_j=\mathbf a^T \mathbf x_j$ are also similar. 
We can further find $d$ more projection vectors so that the data dimension $D$ can be reduced to $d$.

\subsection{Grassmann Manifold and its Distances}

\begin{Def}
\textbf{(Grassmann Manifold)} \cite{AbsilMahonySepulchre2008} The Grassmann manifold, denoted by $\mathcal{G}(p, d)$, consists of  all the $p$-dimensional subspaces embedded in $d$-dimensional Euclidean space  $\mathbb{R}^d$ ($0\leq p \leq d$).
\end{Def}

For example, when $p=0$, the Grassmann manifold becomes the Euclidean space itself. When $p=1$, the Grassmann manifold consists of all the lines passing through the origin in $\mathbb{R}^d$.

As Grassmann manifold is abstract, there are a number of ways to realize it. One convenient way is to represent the manifold by the equivalent classes of all the thin-tall orthogonal matrices under the orthogonal group $\mathcal{O}(p)$ of order $p$. Hence we have the following matrix representation,
\begin{equation}
\begin{aligned}
\mathcal{G}(p,d) = \{\mathbf X \in \mathbb{R}^{d\times p}: \mathbf X^T\mathbf X = \mathbf I_p\} / \mathcal{O}(p).
\end{aligned}
\end{equation}
We refer a point on Grassmann manifold as to an equivalent class of all the thin-tall orthogonal matrices in $\mathbb{R}^{d\times p}$, anyone in which can be converted to the other by a $p\times p$ orthogonal matrix.


There are two popular methods to measure the distance on Grassmann manifold.
One is to define consistent metrics in tangent spaces to make Grassmann manifold a Riemannian manifold.
Another is to embed the Grassmann manifold into symmetric matrices space where the Euclidean metric is available. The later one is easier and more effective in practice, therefore, we use the Embedding distance in this paper.

\begin{Def}
\textbf{(Embedding Distance)} \cite{HarandiSandersonShenLovell2013} Given Grassmann points $\mathbf X_1$ and $\mathbf X_2$, Grassmann manifold can be embedded into symmetric matrices space as,
\begin{equation}
\label{Grassmann2Sym_mapping}
  \Pi : \mathcal{G}(p,d) \rightarrow \text{Sym}(d), \ \ \ \Pi(\mathbf X)=\mathbf X\mathbf X^T,
\end{equation}
and the corresponding distance on Grassmann manifold can be defined as,
\begin{equation}
\label{EmbeddingDistance}
 \text{dist}^2_g(\mathbf X_1,\mathbf X_2) = \frac12\|\Pi(\mathbf X_1)-\Pi(\mathbf X_2)\|^2_F. 
\end{equation}
\end{Def}


\section{The Proposed Method}

In this section, we propose an unsupervised DR method for Grassmann manifold that maps a high-dimensional Grassmann point 
$\mathbf X_i \in \mathcal{G}(p,D)$ to a point in a 
relative low-dimensional Grassmann manifold $\mathcal{G}(p,d), \ D > d$. The mapping $\mathcal{G}(p,D) \rightarrow \mathcal{G}(p,d)$ to be learned is defined as,
\begin{equation}\label{Projection_1}
\begin{aligned}
\mathbf Y_i = \mathbf A^T \mathbf X_i,
\end{aligned}
\end{equation}
where $\mathbf A \in \mathbb{R}^{D\times d}$. To make sure that $\mathbf Y_i \in \mathbb{R}^{d\times p}$ is well-defined as the representative of the mapped Grassmann point on lower dimension manifold, we need impose some conditions.
Obviously, the projected data $\mathbf Y_i$ is not an orthogonal matrix, disqualified as a representative of a Grassmann point. To solve this problem, we perform QR decomposition on matrix $\mathbf Y_i$ as follows \cite{HuangWangShanChen2015},
\begin{equation}\label{Projection_2}
\begin{aligned}
&\mathbf Y_i = \mathbf A^T \mathbf X_i = \mathbf Q_i \mathbf R_i \\
\Rightarrow \ \ &\mathbf Q_i = \mathbf A^T (\mathbf X_i \mathbf R_i^{-1}) = \mathbf A^T \widetilde{\mathbf X}_i, \\
\end{aligned}
\end{equation}
where $\mathbf Q_i \in \mathbb{R}^{d\times p}$ is an orthogonal matrix, $\mathbf R_i \in \mathbb{R}^{p\times p}$ is an invertible upper-triangular matrix, and $\widetilde{\mathbf X}_i = \mathbf X_i\mathbf R_i^{-1} \in \mathbb{R}^{D\times p}$ denotes the normalized $\mathbf X_i$. As both $\mathbf Y_i$ and $\mathbf Q_i$ generate the same (columns) subspace, the orthogonal matrix $\mathbf Q_i$ (or $\mathbf A^T \widetilde{\mathbf X}_i$) can be used as the representative of the low-dimensional Grassmann point mapped from $\mathbf X_i$.

\subsection{LPP for Grassmann manifold (GLPP)}
The term $\left( \mathbf a^T\mathbf x_i - \mathbf a^T\mathbf x_j\right)^2$ in LPP objective function \eqref{LPP_objective} means the distance between the projected data $\mathbf a^T \mathbf x_i$ and $\mathbf a^T \mathbf x_j$; therefore, it is natural for us to reformulate the classic LPP objective function on Grassmann manifold as follows,
\begin{equation}\label{GLPP_Objective1}
\begin{aligned}
 &\min\limits_{\mathbf A}\sum\limits_{ij}^N\text{dist}_g^2(\mathbf Q_i,\mathbf Q_j)\cdot w_{ij} =\sum\limits_{ij}^N\text{dist}_g^2(\mathbf A^T \widetilde{\mathbf X}_i, \mathbf A^T \widetilde{\mathbf X}_j)\cdot w_{ij} \\
\end{aligned}
\end{equation}
where $w_{ij}$ reflects the similarity between original Grassmann points $\mathbf X_i$ and $\mathbf X_j$, and  the distance $\text{dist}_g(\cdot)$ is chosen as the Embedding distance \eqref{EmbeddingDistance}. Hence 
\begin{equation*}\label{GLPP_Objective2}
\begin{aligned}
\text{dist}_g^2(\mathbf A^T \widetilde{\mathbf X}_i, \mathbf A^T \widetilde{\mathbf X}_j) &= \|\mathbf A^T\widetilde{\mathbf X}_i\widetilde{\mathbf X}_i^T\mathbf A - \mathbf A^T\widetilde{\mathbf X}_j\widetilde{\mathbf X}_j^T\mathbf A\|_F^2 \\
&= \|\mathbf A^T \mathbf G_{ij} \mathbf A\|_F^2,
\end{aligned}
\end{equation*}
where $\mathbf G_{ij} = \widetilde{\mathbf X}_i\widetilde{\mathbf X}_i^T - \widetilde{\mathbf X}_j\widetilde{\mathbf X}_j^T$, which is a symmetric matrix of size $D \times D$. Thus, the objective function \eqref{GLPP_Objective1} can be re-written as, termed as GLPP,

\begin{equation}\label{GLPP_Objective2}
\begin{aligned}
\min\limits_{\mathbf A} \sum\limits^N_{i,j=1} \|\mathbf A^T \mathbf G_{ij} \mathbf A\|_F^2\cdot w_{ij}.
\end{aligned}
\end{equation}

The next issue is how to construct the adjacency graph $\mathbf W$ from the original Grassmann points. We extend the Euclidean graph $\mathbf W$ onto Grassmann manifold as follows,

\begin{Def}[Graph $\mathbf W$ on Grassmann manifold] Given a set of Grassmann points $\{\mathbf X_1, ..., \mathbf X_N\}$, we define the graph as
\begin{equation}\label{GrassmannGraph}
\begin{aligned}
w_{ij} &= e^{-\text{dist}_g^2(\mathbf X_i,\mathbf X_j)} \\
\end{aligned}
\end{equation}
where $w_{ij}$ denotes the similarity of Grassmann points $\mathbf X_i$ and $\mathbf X_j$.
\end{Def}

In this definition,  we may set $\text{dist}_g(\mathbf X_i,\mathbf X_j)$ to any one valid Grassmann distance. We select the Embedding distance in our experiments.

\subsection{GLPP with Normalized Constraint}

Without any constraints on $\mathbf A$, we may have a trivial solution from problem \eqref{GLPP_Objective2}. To introduce an appropriate constraint,
we have to firstly define some necessary notations. We split the normalized Grassmann point $\widetilde{\mathbf X}_i \in R^{D\times p}$ and the projected matrix $\mathbf Q_i \in R^{d\times p}$ in \eqref{Projection_2} into their components
\[
\mathbf Q_i = [\mathbf q_{i1}, ..., \mathbf q_{ip}] = [\mathbf A^T \widetilde{\mathbf x}_{i1}, ..., \mathbf A^T \widetilde{\mathbf x}_{ip}] = \mathbf A^T \widetilde{\mathbf X}_i
\]
where $\mathbf q_{ij}\in\mathbb{R}^d$ and $\widetilde{\mathbf x}_{ij}\in\mathbb{R}^D$ with $j=1, 2, ..., p$. For each $j$ ($1\leq j\leq p$), define matrix
\[
\mathbf Q^j = [\mathbf q_{1j}, \mathbf q_{2j}, ..., \mathbf q_{Nj}] \in\mathbb{R}^{d\times N}
\]
and
\[
\widetilde{\mathbf X}^j = [\widetilde{\mathbf x}_{1j}, \widetilde{\mathbf x}_{2j}, ..., \widetilde{\mathbf x}_{Nj}] \in\mathbb{R}^{D\times N}.
\]
That is, from all $N$ normalized Grassmann points $\mathbf Q_i$ (or all $N$ normalized Grassmann points $\widetilde{\mathbf X}_i$), we pick their $j$-th column and stack them together. 
Then, it is easy to check that
\begin{equation*}
\mathbf Q^j = \mathbf A^T \widetilde{\mathbf X}^j.
\end{equation*}
For this particularly organized matrix $\mathbf Q^j$, considering the constraint condition similar to formula \eqref{LPP_Constrain},
\begin{align*}
\text{tr}(\mathbf Q^j \mathbf D\mathbf Q^{jT}) =\text{tr}(\mathbf D\mathbf Q^{jT}\mathbf Q^j) = \text{tr}\left(\mathbf D \widetilde{\mathbf X}^{jT}\mathbf A\mathbf A^T\widetilde{\mathbf X}^j\right).
\end{align*}
Hence, one possible overall constraint can be defined as
\[
\sum^p_{j=1}\text{tr}\left(\mathbf D \widetilde{\mathbf X}^{jT}\mathbf A\mathbf A^T\widetilde{\mathbf X}^j\right) = 1.
\]
Rather than using the notation $\widetilde{\mathbf X}^j$, we can further simplify it into a form by using original normalized Grassmann points $\widetilde{\mathbf X}_i$.  A long algebraic manipulation can prove that
\begin{align*}
&\sum^p_{j=1}\text{tr}\left(\mathbf D \widetilde{\mathbf X}^{jT}\mathbf A\mathbf A^T\widetilde{\mathbf X}^j\right)
=\text{tr}\left(\mathbf A^T \left(\sum^N_{i=1}d_{ii}\widetilde{\mathbf X}_i\widetilde{\mathbf X}^T_i \right)\mathbf A \right).
\end{align*}
Hence, we add the following constraint condition
\[
\text{tr}\left(\mathbf A^T \left(\sum^N_{i=1}d_{ii}\widetilde{\mathbf X}_i\widetilde{\mathbf X}^T_i \right)\mathbf A \right) = 1.
\]
Define $\mathbf H = \sum^N_{i=1}d_{ii}\widetilde{\mathbf X}_i\widetilde{\mathbf X}^T_i$, then the final constraint condition can be written as,
\begin{equation}\label{Constraint_final}
\begin{aligned}
\text{tr}(\mathbf A^T \mathbf H \mathbf A) = 1.
\end{aligned}
\end{equation}
Combining the objective function \eqref{GLPP_Objective2} and constraint condition \eqref{Constraint_final}, we get the overall GLPP model,
\begin{equation}\label{GLPP_final}
\begin{aligned}
\min\limits_{\mathbf A} \sum\limits_{i,j=1}^{N}\|\mathbf A^T\mathbf G_{ij}\mathbf A\|_F^2\cdot w_{ij} \ \ \text{s.t.} \ \ \text{tr}(\mathbf A^T\mathbf H \mathbf A) = 1
\end{aligned}
\end{equation}

\begin{algorithm}
\renewcommand{\algorithmicrequire}{\textbf{Input:}}
\renewcommand\algorithmicensure {\textbf{Output:} }
\caption{ LPP for Grassmann manifold.}\label{wholeAlg1a}
\begin{algorithmic}[1]
\REQUIRE Grassmann points $\{\mathbf X_i\}_{i=1}^N$, \ $\mathbf X_i\in \mathcal{G}(p,D)$.  \\
\ENSURE  The mapping $\mathbf A \in R^{D\times d}$. ~~\\
\STATE   Initialize: Set the parameter $\mathbf A^{(0)} = \begin{bmatrix} I_{d\times d}\\ \text{random elements} \end{bmatrix}$. \\
\STATE   Calculate graph $\mathbf W$ of original Grassmann data $\mathbf X_i$ according to the formula \eqref{GrassmannGraph}.
\WHILE   {not converged}
\STATE Normalize $\mathbf X_i$ by using $\widetilde{\mathbf X}_i^{(k+1)}=\mathbf X_i^{(k)}\mathbf R_i^{{(k)}^{-1}}$ where $\mathbf A^{(k)^T}\mathbf X_i^{(k)} = \mathbf Q_i^{(k)}\mathbf R_i^{(k)}$. \\
\STATE Compute $\mathbf G_{ij}^{(k+1)} = \widetilde{\mathbf X}_i^{(k+1)}\widetilde{\mathbf X}_i^{(k+1)^T}-\widetilde{\mathbf X}_j^{(k+1)}\widetilde{\mathbf X}_j^{(k+1)^T}$ and $\mathbf H^{(k+1)} = \sum\limits_{i=1}^{N}d_{ii}\widetilde{\mathbf X}_i^{(k+1)}\widetilde{\mathbf X}_i^{(k+1)^T}$. \\
\STATE Optimize $\mathbf A^{(k+1)}$ in equation \eqref{GLPP_final} by solving an generalized eigenvalue problem.\\
\ENDWHILE
\end{algorithmic}
\end{algorithm}

In next section, we propose a simplified way to solve problem \eqref{GLPP_final} which is quite different from most Riemannian manifold based optimization algorithms such as in the Riemannian Conjugate Gradient (RCG) toolbox.

\section{Optimization}
In this section, we provide an iteration solution to solve the optimization problems \eqref{GLPP_final}. First we write the  cost function  as follows
\[
f(\mathbf A) = \sum^N_{i,j=1}\text{tr}\left( \mathbf A^T \mathbf G_{ij} \mathbf A\mathbf A^T\mathbf G_{ij} \mathbf A\right)\cdot w_{ij}.
\]
For ease, we redefine a new objective function $f_k$ in the $k-$th iteration by using the last step $\mathbf A^{(k-1)}$ as the following way,
\begin{equation}
\begin{aligned}
f_k(\mathbf A) &= \sum^N_{i,j=1}w_{ij}\cdot\text{tr}\left( \mathbf A^T \mathbf G_{ij} \mathbf A^{(k-1)}\mathbf A^{(k-1)T}\mathbf G_{ij} \mathbf A\right) \\
 &= \text{tr}\left(\mathbf A^T \sum^N_{i,j=1} w_{ij}\mathbf G_{ij} \mathbf A^{(k-1)}\mathbf A^{(k-1)T}\mathbf G_{ij} \mathbf A\right).
\end{aligned}
\end{equation}
Denoting
\[
\mathbf J = \sum^N_{i,j=1}w_{ij}\mathbf G_{ij} \mathbf A^{(k-1)}\mathbf A^{(k-1)T}\mathbf G_{ij},
\]
where $\mathbf G_{ij}$ is calculated according to $\mathbf A^{(k-1)}$ through both $\widetilde{\mathbf X}_i$ and $\widetilde{\mathbf X}_j$.
Then the simplified version of problem \eqref{GLPP_final} becomes
\begin{equation}\label{Sec4:objective}
\begin{aligned}
\min_{\mathbf A} \text{tr}(\mathbf A^T\mathbf J\mathbf A),\;\;\; \text{s.t.}\;\;\; \text{tr}(\mathbf A^T\mathbf H\mathbf A) = 1.
\end{aligned}
\end{equation}
The Lagrangian function of \eqref{Sec4:objective} is given by

\begin{equation}
\begin{aligned}
 \text{tr}(\mathbf A^T \mathbf J \mathbf A) + \lambda (1 - \text{tr}(\mathbf A^T\mathbf H\mathbf A)),
\end{aligned}
\end{equation}
which can be derived to solve and translated to a generalized eigenvalue problem,
\[
\mathbf J \mathbf a = \lambda \mathbf H\mathbf a.
\]
Obviously, matrices $\mathbf H$ and $\mathbf J$ are symmetrical and positive semi-definite. By performing eigenvalue decomposition on $\mathbf H^{-1}\mathbf J$, the transform matrix $\mathbf A = [\mathbf a_1, ..., \mathbf a_d]\in \mathbb{R}^{D\times d}$ is given by the minimum $d$ eigenvalue solutions to the generalized eigenvalue problem.


We summarize the whole procedures as Algorithm \ref{wholeAlg1a}.


\section{Experiments}

In this section, we evaluate our proposed method GLPP on several classification and clustering tasks, respectively. 

\subsection{Experimental settings}


\begin{table*}
   \centering
   \begin{tabular}{c||ccccccc}
     \hline
              Methods     & Num of samples & Num of clusters & $D$ & $d$ & $r$ & $\mathbf X_i$ & $\mathbf Q_i$  \\
              \hline \hline
              Extended Yale B & 297 & 38 & 400 & 62 & 0.95 & $\mathcal{G}(4,400)$ & $\mathcal{G}(4,62)$  \\
              \hline
              Highway Traffic & 253 & 3 & 576 & 163 & 0.95 & $\mathcal{G}(10,576)$ & $\mathcal{G}(10,83)$ \\
              \hline
              UCF Sport & 150 & 13 & 900 &405 & 0.95 & $\mathcal{G}(20,900)$ & $\mathcal{G}(20,405)$ \\
              \hline
   \end{tabular}\\[2mm]
   \caption{Parameters list. Parameters $d$ and $r$ denote the reduced dimensionality and the remaining energy rate. $\mathbf X_i$ and $\mathbf Q_i$ represent the original high-dimensional Grassmann manifold and the new low-dimensional Grassmann manifold, respectively.}\label{Parametertab}
\end{table*}

\subsubsection{Datasets}

\emph{Extended Yale B dataset}\footnote{http://vision.ucsd.edu/content/yale-face-database} is captured from $38$ subjects and each subject has $64$ front face images in different light directions and illumination conditions. All images are resized into $20\times 20$ pixels. 

\emph{Highway Traffic dataset}\footnote{http://www.svcl.ucsd.edu/projects/traffic/} 
contains 253 video sequences of highway traffic. These sequences are labeled with three levels: 44 clips at heavy level, 45 clips at medium level and 164 clips at light level. Each video sequence has 42 to 52 frames. The video sequences are converted to gray images and each image is normalized to size  $24 \times 24$. 

\emph{UCF sport dataset}\footnote{http://crcv.ucf.edu/data/} 
includes a total of 150 sequences. The collection has a natural pool of actions with a wide range of scenes and viewpoints. 
There are 13 actions in this dataset. 
Each sequence has 22 to 144 frames. We convert these video clips into gray images and each image is resized into $30\times 30$.

\begin{figure}[!h]
\centering
\subfigure[]{ \label{Yalefig} 
\includegraphics[width=0.45\textwidth]{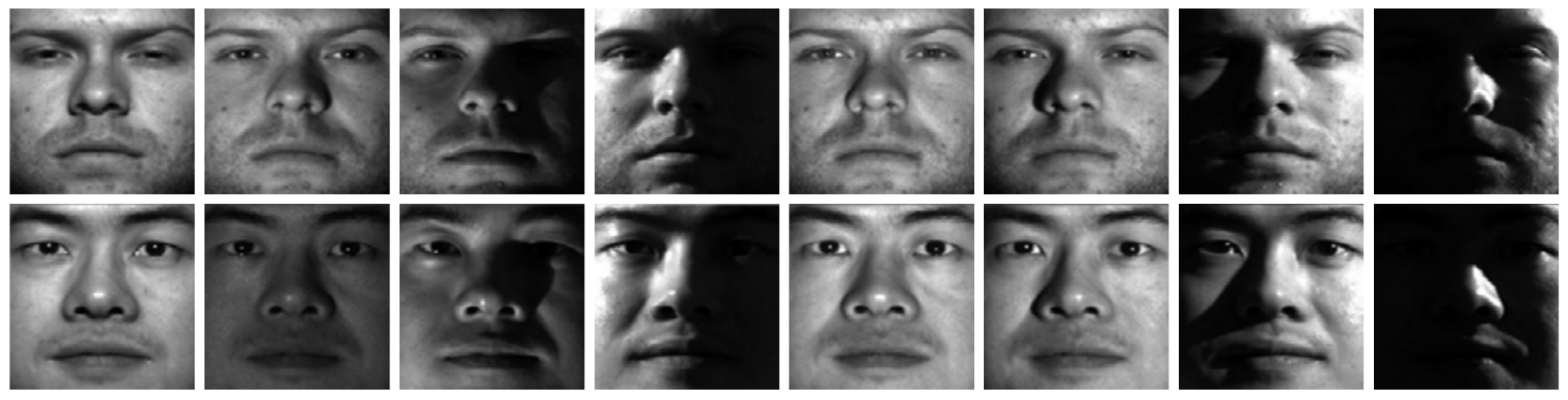}}
\hspace{0.1in}
\subfigure[]{ \label{Trafficfig} 
\includegraphics[width=0.45\textwidth]{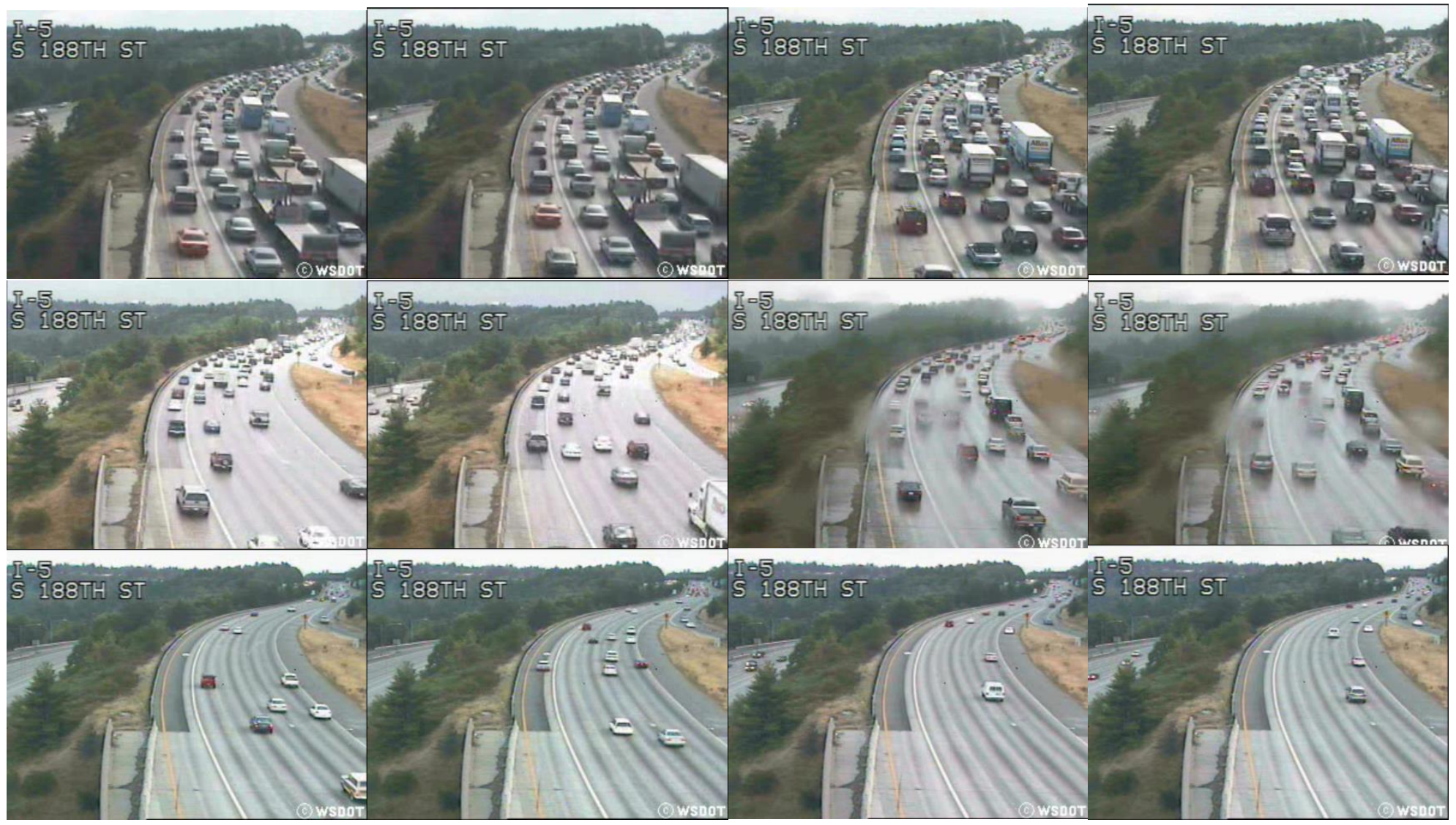}}
\hspace{0.1in}
\subfigure[]{ \label{UCFfig} 
\includegraphics[width=0.45\textwidth]{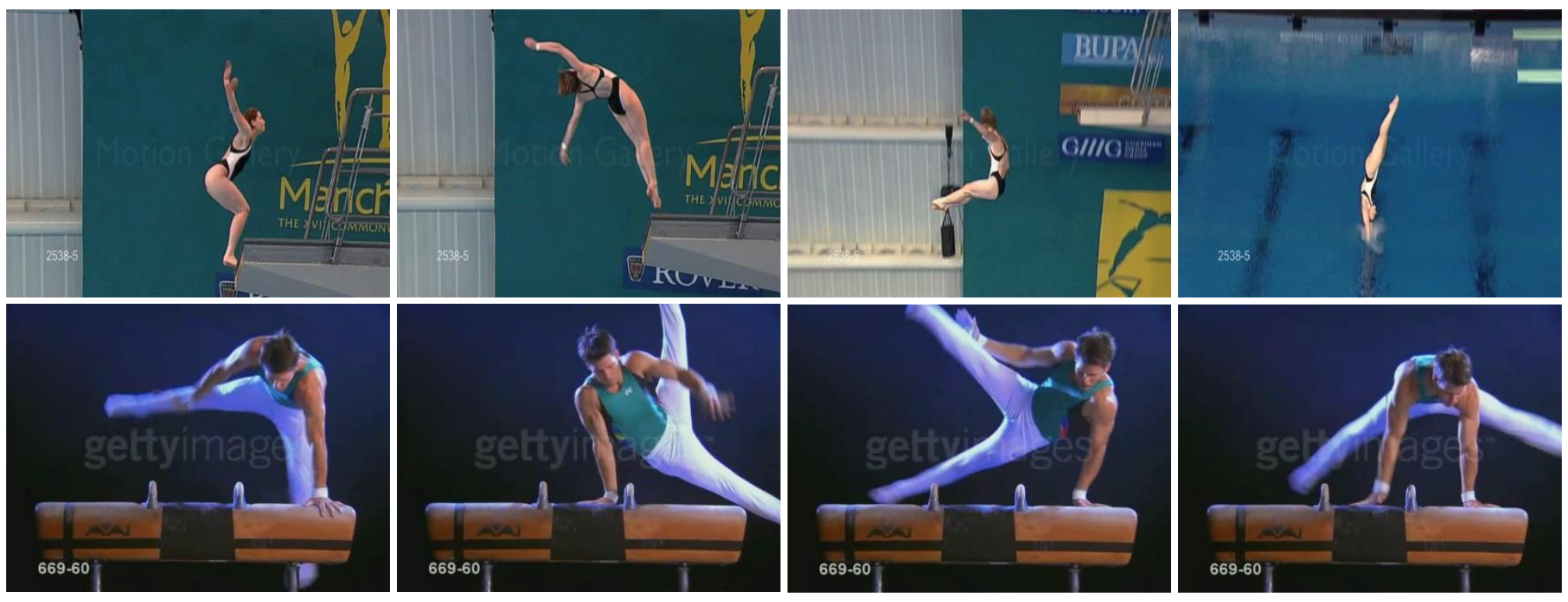}}
\caption{Some samples from different datasets. (a) Extended Yale B dataset. Each row denotes an image set sample which contains $8$ face images captured from different light directions and illuminations; (b) Highway Traffic dataset; (c) UCF sport dataset.} \label{Balletfig}
\end{figure}

Figure \ref{Balletfig} shows some samples from these three datasets.

\begin{table*}
   \centering
   \begin{tabular}{c||cc|cccc}
     \hline
              Evaluation &\multicolumn{2}{c|}{Num of Samples} & \multicolumn{4}{c}{ACC} \\
              Methods     & Training & Testing &  GKNN & GKNN-GLPP & GDL & GDL-GLPP \\
              \hline \hline
              Dataset              & \multicolumn{6}{c}{Extended Yale B} \\
              \hline
              38 sub  & 221 & 76 &  94.74 &  \textbf{1} & 1 & \textbf{1} \\
              \hline \hline
              Dataset              & \multicolumn{6}{c}{Highway Traffic} \\
              \hline
              3 sub & 192 & 60 &  70.00 & \textbf{76.67} & 65.00 & \textbf{70.00}\\
              \hline \hline
              Dataset              & \multicolumn{6}{c}{UCF Sport} \\
              \hline
              13 sub  & 124 & 26 &  53.85 &  \textbf{61.54} & 61.54 & \textbf{65.38}\\
              \hline \hline
   \end{tabular}\\[2mm]
      \caption{Classification results (in $\%$) on different datasets. We also list the number of samples in the first two columns. The figures in boldface give the best performance among all the compared methods.}\label{Classificationtab}
\end{table*}

\subsubsection{Parameters and Evaluation}

The reduced dimension $d$ is the most important parameter for DR algorithms. Like PCA, we define $d$ by the cumulative energy of the eigenvectors, i.e. given the remaining energy rate $r \; (0 < r < 1)$, $d$ is defined as follows,

\[
d = \arg\min\{ d^*\in \mathbb N: \sum\limits_{i=1}^{d^*}\sigma_i \geq r \sum\limits_{i=1}^{D} \sigma_i\},
\]
where $\sigma_i$ is the $i$-th largest eigenvalue of $\mathbf P\mathbf P^T$, in which we stack all the Grassmann points $\mathbf P = [\mathbf X_1 ; ... ; \mathbf X_N]$.
However, for different datasets and applications, it is difficult to set a proper $r$ uniformly. 
For simplification and fairness, here we set $r=0.95$ in all our experiments.

\begin{figure}
    \begin{center}
    \includegraphics[width=0.45\textwidth]{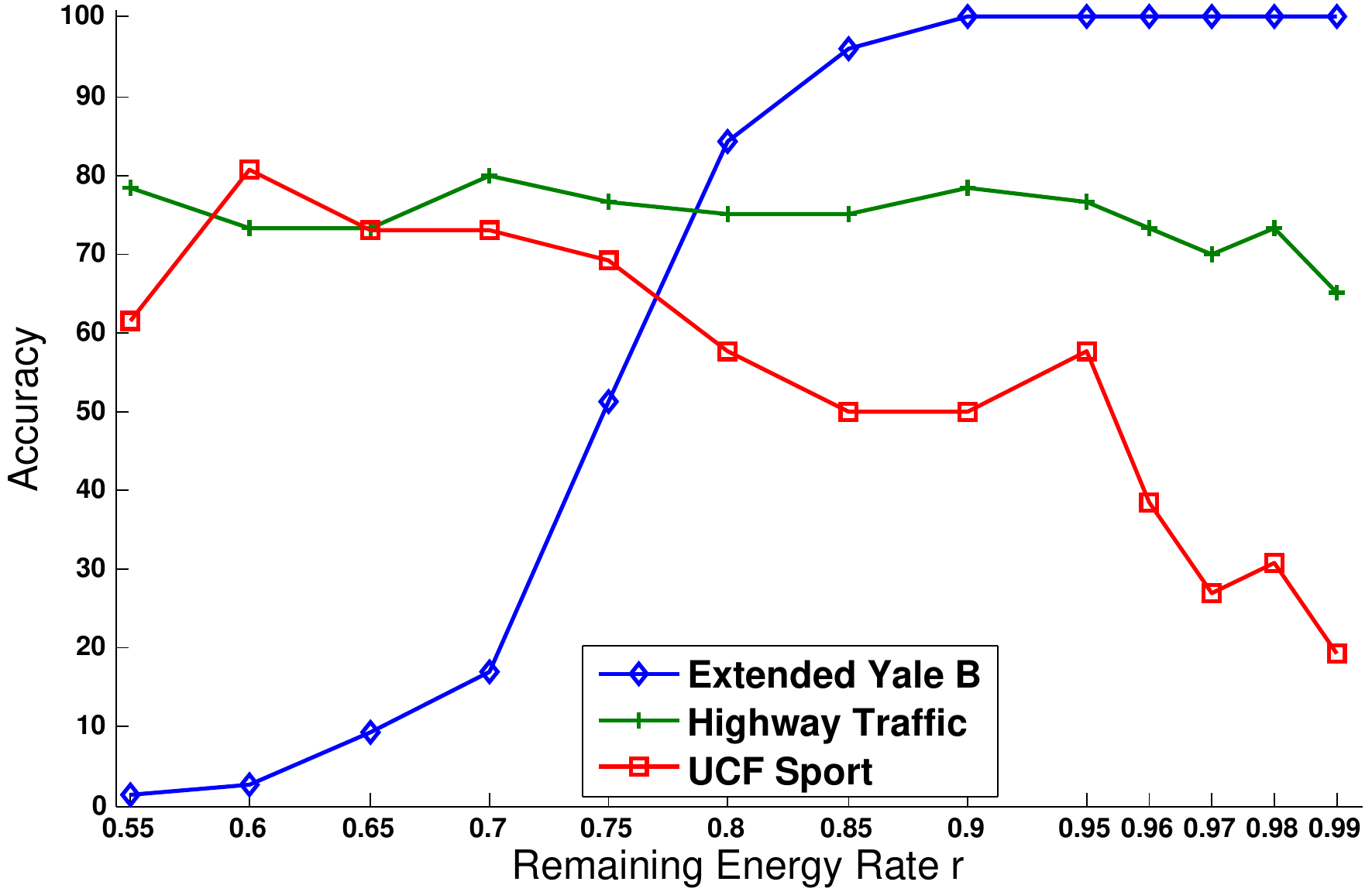}
    \end{center}
    \caption{The experimental results of GKNN-GLPP corresponding to the remaining energy rate $r$ from $0.55$ to $0.99$.}\label{UCFfig}
\end{figure}

The performance of different algorithms is evaluated by Accuracy (ACC) and we also add Normalized Mutual Information (NMI) \cite{Kvalseth1987} as an additional evaluation method for clustering algorithms.
ACC reflects the percentage of correctly labeled samples, while NMI calculates the mutual dependence of the predicted clustering and the ground-truth partitions.

For the sake of saving space, we list all experimental parameters in Table \ref{Parametertab}. All the algorithms are coded in Matlab 2014a and implemented on an Intel Core i7-4600M 2.9GHz CPU machine with 8G RAM.

\subsection{Video/Imageset Classification}

We firstly evaluate the performance of GLPP on classification task, and we use $K$ Nearest Neighbor on Grassmann manifold algorithm (GKNN) and Dictionary Learning on Grassmann manifold (GDL) \cite{HarandiSandersonShenLovell2013} as baselines,

\begin{itemize}
  \item \textbf{GKNN:}   KNN classifier  based on the Embedding distance on high-dimensional Grassmann manifold;
  \item \textbf{GKNN-GLPP:} KNN classifier on low-dimensional Grassmann manifold obtained by the proposed method;
  \item \textbf{GDL:} GDL on high-dimensional Grassmann manifold;
  \item \textbf{GDL-GLPP:} GDL on low-dimensional Grassmann manifold.
\end{itemize}

Human facial recognition is one of the hottest topics in computer vision and pattern recognize area. Affected by various factors, i.e., expression, illumination conditions and light directions, algorithms based on individual faces do not achieve great experimental performance. Therefore, we test our proposed method GLPP on classic Extended Yale B dataset.
We wish to inspect the proposed method on a practical application in complex environment; therefore we pick the Highway Traffic video dataset which contains various weather conditions, such as sunny, cloudy and rainy. UCF sport dataset which contains more variations on scenes and viewpoints can be used to examine the robustness of the proposed methods in noised scenarios. In our experiments, each video clip is regarded as an imageset.

To be fair, we set $K=5$ for GKNN algorithm in all three experiments, and the number of training and testing samples are listed in the first two columns in Table \ref{Classificationtab}, while other parameters can be found in Table \ref{Parametertab} (i.e., $D$, $d$ and $r$).

Experimental results for classification tasks are shown in Table \ref{Classificationtab}. Obviously, the experimental accuracy of GLPP-based algorithms is at least 5 percent higher than the corresponding compared methods in most cases. We distribute it to LPP is less sensitive to outliers since LPP is derived by preserving local information. The experimental results also demonstrate that the low-dimensional Grassmann points generated by our proposed method reflect more discrimination than on the original Grassmann manifold.

How to infer the intrinsic dimensionality from high-dimensional data still is a challenging problem. The intrinsic dimensionality relies heavily on practical applications and datasets.  In our method, the reduced dimensionality $d$ is determined by the remaining energy rate $r$.
Figure \ref{UCFfig} shows that there exist different optimal $r$ or $d$ for different datasets. Extended Yale B dataset contains much rich information (e.g., face contour, texture, expression, illustration conditions and light directions) which has strong impacts on face recognition accuracy. When the reduced dimensionality $d$ is less than the intrinsic dimensionality, the data in reduced dimensionality may lose some useful discriminative information. Therefore, the accuracy increases with larger reduced dimensionality in a certain range, e.g., the remaining energy rate $r$ from $0.6$ to $0.95$ for Extended Yale B dataset. We find the optimal value is $r=0.96$. For the Traffic and UCF datasets, the simple or static backgrounds occupy main area of images. The foreground, e.g. car action and human action, is more valuable information for classification. When the optimal reduced dimensionality $d$ is achieved at relative small $r$, here $0.7$ and $0.8$, the data in reduced dimensionality actually contain the ¡°right¡± information for Traffic and UCF data. In other words, when the reduced dimensionality $d$ is getting larger ($>$ the intrinsic dimensionality), the information from the background may lead to negative influence for the accuracy.

\subsection{Video/Imageset Clustering}

To further verify the performance of GLPP, we apply it on clustering tasks, and select K-means on Grassmann manifold (GKM) \cite{TuragaVeeraraghavanSrivastavaChellappa2011} as the compared mathod, 

\begin{itemize}
  \item \textbf{GKM} : K-means based on the Embedding distance on high-dimensional Grassmann manifold.
  \item \textbf{GKM-GLPP}: K-means on low-dimensional Grassmann manifold obtained by our proposed method.
\end{itemize}

Table \ref{ACCtab} shows ACC and NMI values for all algorithms.
Clearly, after drastically reducing dimensionality from $D$ to $d$ (see Table \ref{Parametertab})  by our proposed method, the new low-dimensional Grassmann manifold still maintain fairly higher accuracy than the original high-dimensional Grassmann manifold for all algorithms, which attests that our proposed DR scheme significantly boosts the performance of GKM.

\begin{table}
   \centering
   \small
   \begin{tabular}{c||cc|cc}
     \hline
              Evaluation  & \multicolumn{2}{c|}{ACC} & \multicolumn{2}{c}{NMI} \\
              Methods     & GKM & GKM-GLPP & GKM & GKM-GLPP  \\
              \hline \hline
              Dataset& \multicolumn{4}{c}{Extended Yale B}  \\
              \hline
              38 sub   & 56.57 & \textbf{80.47} & 76.02 & \textbf{91.08} \\
              \hline \hline
              Dataset& \multicolumn{4}{c}{Highway Traffic}  \\
              \hline
               3 sub     & 64.43 & \textbf{73.52} & 27.13 & \textbf{38.59}\\
              \hline \hline
              Dataset& \multicolumn{4}{c}{UCF Sport}  \\
              \hline
              13 sub     & 50.00 & \textbf{57.33} & 56.54 & \textbf{62.70}\\
              \hline \hline
   \end{tabular}\\[2mm]
   \caption{Clustering results (in $\%$) on different datasets. The figures in boldface give the best performance among all the compared methods.}\label{ACCtab}
\end{table}

\section{Conclusion}
In this paper, we extended the unsupervised LPP algorithm onto Grassmann manifold by learning a projection from the high-dimensional Grassmann manifold into the one in a relative low-dimensional with more discriminative capability, based on the strategy of embedding Grassamnn manifolds onto the space of symmetric matrices. 
The basic idea of LPP is to preserve the local structure of original data in the projected space.
Our proposed model can be  simplified as a basic eigenvalue problem for an easy solution.
Compared with directly using the high-dimensional Grassmann manifold, the experimental results illustrate the effectiveness and superiority of the proposed GLPP on video/imageset classification and clustering tasks.

\section*{Acknowledgements}
The research project is supported by the Australian Research Council (ARC) through the grant DP140102270 and also partially supported by National Natural Science Foundation of China under Grant No. 61390510, 61672071, 61632006, 61370119, Beijing Natural Science Foundation No. 4172003, 4162010, 4152009, Beijing Municipal Science $\&$ Technology Commission No. Z171100000517003, Project of Beijing Municipal Education Commission No. KM201610005033, Funding Project for Academic Human Resources Development in Institutions of Higher Learning Under the Jurisdiction of Beijing Municipality No.IDHT20150504 and Beijing Transportation Industry Science and Technology Project.

\bibliographystyle{named}

\begin{thebibliography}{}

\bibitem[\protect\citeauthoryear{Absil \bgroup \em et al.\egroup
  }{2008}]{AbsilMahonySepulchre2008}
P.~Absil, R.~Mahony, and R.~Sepulchre.
\newblock {\em Optimization {A}lgorithms on {M}atrix {M}anifolds}.
\newblock Princeton University Press, 2008.

\bibitem[\protect\citeauthoryear{Bishop}{2006}]{Bishop2006}
C.~M. Bishop.
\newblock {\em Pattern Recognition and Machine Learning (Information Science
  and Statistics)}.
\newblock Springer-Verlag New York, Inc., Secaucus, NJ, USA, 2006.

\bibitem[\protect\citeauthoryear{Chen \bgroup \em et al.\egroup
  }{2007}]{ChenZhaoKongLuo2007}
S.~Chen, H.~Zhao, M.~Kong, and B.~Luo.
\newblock {2D-LPP}: A two-dimensional extension of locality preserving
  projections.
\newblock {\em Neurocomputing}, 70:912--921, 2007.

\bibitem[\protect\citeauthoryear{Comon}{1994}]{Comon1994}
P.~Comon.
\newblock Independent component analysis, a new concept?
\newblock {\em Signal Processing}, 36(3):287--314, 1994.

\bibitem[\protect\citeauthoryear{Faraki \bgroup \em et al.\egroup
  }{2015}]{FarakiHarandiPorikli2015b}
M.~Faraki, M.~T. Harandi, and F.~Porikli.
\newblock More about {VLAD}: A leap from {E}uclidean to {R}iemannian manifolds.
\newblock In {\em CVPR},
  2015.

\bibitem[\protect\citeauthoryear{Hamm and Lee}{2008}]{HammLee2008}
J.~Hamm and D.~Lee.
\newblock {G}rassmann discriminant analysis: a unifying view on sub-space-based
  learning.
\newblock In {\em ICML}, 2008.

\bibitem[\protect\citeauthoryear{Harandi \bgroup \em et al.\egroup
  }{2011}]{HarandiSandersonShiraziLovell2011}
M.~T. Harandi, C.~Sanderson, S.~A. Shirazi, and B.~C. Lovell.
\newblock Graph embedding discriminant analysis on {G}rassmannian manifolds for
  improved image set matching.
\newblock In {\em CVPR}, 2011.

\bibitem[\protect\citeauthoryear{Harandi \bgroup \em et al.\egroup
  }{2013}]{HarandiSandersonShenLovell2013}
M.~T. Harandi, C.~Sanderson, C.~Shen, and B.~Lovell.
\newblock Dictionary learning and sparse coding on {G}rassmann manifolds: {A}n
  extrinsic solution.
\newblock In {\em ICCV}, 2013.

\bibitem[\protect\citeauthoryear{Harandi \bgroup \em et al.\egroup
  }{2014}]{HarandiSalzmannHartley2014}
M.~T. Harandi, M.~Salzmann, and R.~Hartley.
\newblock From manifold to manifold: Geometry-aware dimensionality reduction
  for {SPD} matrices.
\newblock In {\em ECCV}, 2014.

\bibitem[\protect\citeauthoryear{Harandi \bgroup \em et al.\egroup
  }{2015}]{HarandiHartleyLovellSanderson2015}
M.~T. Harandi, R.~Hartley, B.~Lovell, and C.~Sanderson.
\newblock Sparse coding on symmetric positive definite manifolds using
  {B}regman divergences.
\newblock {\em IEEE TNNLS},
  27(6):1294--1306, 2015.

\bibitem[\protect\citeauthoryear{He and Niyogi}{2003}]{HeNiyogi2003}
X.~He and P.~Niyogi.
\newblock Locality preserving projections.
\newblock In {\em NIPS}, 2003.

\bibitem[\protect\citeauthoryear{He \bgroup \em et al.\egroup
  }{2005}]{HeCaiYanZhang2005}
X.~He, D.~Cai, S.~Yan, and H.~Zhang.
\newblock Neighborhood preserving embedding.
\newblock In {\em ICCV}, 2005.

\bibitem[\protect\citeauthoryear{Huang \bgroup \em et al.\egroup
  }{2014}]{HuangWangShanChen2014}
Z.~Huang, R.~Wang, S.~Shan, and X.~Chen.
\newblock Learning {E}uclidean-to-{R}iemannian metric for point-to-set
  classification.
\newblock In {\em CVPR},
  2014.

\bibitem[\protect\citeauthoryear{Huang \bgroup \em et al.\egroup
  }{2015}]{HuangWangShanChen2015}
Z.~Huang, R.~Wang, S.~Shan, and X.~Chen.
\newblock Projection metric learning on {G}rassmann manifold with application
  to video based face recognition.
\newblock In {\em CVPR},
  2015.

\bibitem[\protect\citeauthoryear{Jayasumana \bgroup \em et al.\egroup
  }{2015}]{JayasumanaHartleySalzmannLiHarandi2015}
S.~Jayasumana, R.~Hartley, M.~Salzmann, H.~Li, and M.~Harandi.
\newblock Kernel methods on {R}iemannian manifolds with {G}aussian {RBF}
  kernels.
\newblock {\em IEEE PAMI},
  37(12):2464--2477, 2015.

\bibitem[\protect\citeauthoryear{Ju \bgroup \em et al.\egroup
  }{2015}]{JuSunGaoHuYin2015}
F.~Ju, Y.~Sun, J.~Gao, Y.~Hu, and B.~Yin.
\newblock Image outlier detection and feature extraction via {L}1-norm based
  2{D} probabilistic {PCA}.
\newblock {\em IEEE TIP}, 24(12):4834--4846, 2015.

\bibitem[\protect\citeauthoryear{Kvalseth}{1987}]{Kvalseth1987}
T.~O. Kvalseth.
\newblock Entropy and correlation : Some comments.
\newblock {\em IEEE TSMC Part C},
  17(3):517--519, 1987.

\bibitem[\protect\citeauthoryear{Luo \bgroup \em et al.\egroup
  }{2016}]{LuoNieChangYang2016}
M.~Luo, F.~Nie, X.~Chang, Y.~Yang, A.~Hauptmann, and Q.~Zheng.
\newblock Avoiding optimal mean robust {PCA}/{2DPCA} with non-greedy $\ell_1$-norm
  maximization.
\newblock In {\em IJCAI}, 2016.

\bibitem[\protect\citeauthoryear{Nguyen \bgroup \em et al.\egroup
  }{2008}]{NguyenLiuVenkatesh2008}
N.~Nguyen, W.~Liu, and S.~Venkatesh.
\newblock Ridge regression for two dimensional locality preserving projection.
\newblock In {\em ICPR}, 2008.

\bibitem[\protect\citeauthoryear{Pennec \bgroup \em et al.\egroup
  }{2006}]{PennecFillardAyache2006}
X.~Pennec, P.~Fillard, and N.~Ayache.
\newblock A {R}iemannian framework for tensor computing.
\newblock {\em IJCV}, 66(1):41--66, 2006.

\bibitem[\protect\citeauthoryear{Sun \bgroup \em et al.\egroup
  }{2010}]{SunCeranYe2010}
L.~Sun, B.~Ceran, and J.~Ye.
\newblock A scalable two-stage approach for a class of dimensionality reduction
  techniques.
\newblock In {\em KDD}, 2010.

\bibitem[\protect\citeauthoryear{Turaga \bgroup \em et al.\egroup
  }{2011}]{TuragaVeeraraghavanSrivastavaChellappa2011}
P.~Turaga, A.~Veeraraghavan, A.~Srivastava, and R.~Chellappa.
\newblock Statistical computations on {G}rassmann and {S}tiefel manifolds for
  image and video-based recognition.
\newblock {\em IEEE TPAMI},
  33(11):2273--2286, 2011.

\bibitem[\protect\citeauthoryear{Wang and Gao}{2016}]{WangGao2016}
Q.~Wang and Q.~Gao.
\newblock Robust 2{DPCA} and its application.
\newblock In {\em CVPR},
  2016.

\bibitem[\protect\citeauthoryear{Wang \bgroup \em et al.\egroup
  }{2008}]{WangChenHuLuo2008}
H.~Wang, S.~Chen, Z.~Hu, and B.~Luo.
\newblock Probabilistic two-dimensional principal component analysis and its
  mixture model for face recognition.
\newblock {\em Neural Computing and Applications}, 17(5-6):541--547, 2008.

\bibitem[\protect\citeauthoryear{Wang \bgroup \em et al.\egroup
  }{2012}]{WangGuoDavisDai2012}
R.~Wang, H.~Guo, L.~Davis, and Q.~Dai.
\newblock Covariance discriminative learning: A natural and efficient approach
  to image set classification.
\newblock In {\em CVPR}, 2012.

\bibitem[\protect\citeauthoryear{Wang \bgroup \em et al.\egroup
  }{2016}]{WangHuGaoSunYin2016}
B.~Wang, Y.~Hu, J.~Gao, Y.~Sun, and B.~Yin.
\newblock Product {G}rassmann manifold representation and its {LRR} models.
\newblock In {\em AAAI}, 2016.

\bibitem[\protect\citeauthoryear{Xie \bgroup \em et al.\egroup
  }{2016}]{XieTaoWei2016}
L.~Xie, D.~Tao, and H.~Wei.
\newblock Multi-view exclusive unsupervised dimension reduction for video-based
  facial expression recognition.
\newblock In {\em IJCAI}, 2016.

\bibitem[\protect\citeauthoryear{Xu \bgroup \em et al.\egroup
  }{2009}]{XuFengZhao2009}
Y.~Xu, G.~Feng, and Y.~Zhao.
\newblock One improvement to two-dimensional locality preserving projection
  method for use with face recognition.
\newblock {\em Neurocomputing}, 73(2009):245--249, 2009.

\bibitem[\protect\citeauthoryear{Yang \bgroup \em et al.\egroup
  }{2004}]{YangZhangFrangiYang2004}
J.~Yang, D.~Zhang, A.~F. Frangi, and J.~Yang.
\newblock Two dimensional {PCA}: A new approach to appearance-based face
  representation and recognition.
\newblock {\em IEEE TPAMI},
  26(1):131--137, 2004.

\bibitem[\protect\citeauthoryear{Yu \bgroup \em et al.\egroup
  }{2008}]{YuBiYe2008}
S.~Yu, J.~Bi, and J.~Ye.
\newblock Matrix-variate factor analysis and its applications.
\newblock In {\em In KDD Workshop}, 2008.

\end{thebibliography}

\end{document}